\documentclass[letterpaper, 10 pt, conference]{ieeeconf}  

\IEEEoverridecommandlockouts                              

\usepackage{amsmath}
\usepackage{leftidx}
\usepackage{graphics} 
\usepackage{epsfig} 
\usepackage{amsmath} 
\usepackage{amssymb}  
\usepackage[usenames,dvipsnames,svgnames,table]{xcolor} 
\usepackage{bm}
\usepackage[percent]{overpic}
\usepackage{lipsum}
\usepackage[font=footnotesize]{subfig}
\usepackage{graphicx}
\usepackage{blindtext}
\usepackage{caption}
\usepackage[normalem]{ulem}
\usepackage{url}
\usepackage{multirow}
\usepackage{hyperref}
\usepackage[switch]{lineno}
\usepackage{footmisc}

\usepackage{textcomp}
\usepackage{tikz}
\usetikzlibrary{shapes,arrows}
\usetikzlibrary{calc}
\usetikzlibrary{positioning}
\tikzset{%
  block/.style    = {draw, thick, rectangle, minimum height = 3em,
    minimum width = 3em},
  input/.style    = {coordinate}, 
  output/.style   = {coordinate} 
}

\newcommand{\vc}[1]{\mathbf{#1}} 					
\newcommand{\cv}[0]{\mathbf{c}} 					
 
\DeclareMathOperator*{\st}{\mathbf{s.t.}\,}					
\newcommand{\T}[0]{\ensuremath{\top}}							
\newcommand{\Expect}{{\rm I\kern-.3em E}}				

\newcommand{\vx}[0]{\mathbf{p}}							%
\title{\LARGE \bf SL1M: Sparse L1-norm Minimization for contact planning on uneven terrain
}

\author{Steve Tonneau$^{1}$, Daeun Song$^{2}$, Pierre Fernbach$^{3}$, Nicolas Mansard$^{3}$, Michel Ta\"ix$^{3}$ and Andrea Del Prete$^{4}$
\thanks{$^{1}$IPAB, The University of Edinburgh, Scotland}%
\thanks{$^{2}$Deparment of Computer Science and Engineering, Ewha Womans University, Korea}%
\thanks{$^{3}$LAAS-CNRS / Universit\'e de Toulouse, France}%
\thanks{$^{4}$Industrial Engineering Department, University of Trento, Italy}%
}

\begin{document}

\maketitle
\thispagestyle{empty}
\pagestyle{empty}

\begin{abstract}
One of the main challenges of planning legged locomotion in complex environments is the combinatorial contact selection problem.
Recent contributions propose to use integer variables to represent which contact surface is selected, and then to rely on modern mixed-integer (MI) optimization solvers to handle this combinatorial issue. To reduce the computational cost of MI, we exploit the sparsity properties of L1 norm minimization techniques to relax the contact planning problem into a feasibility linear program. Our approach accounts for kinematic reachability of the center of mass (COM) and of the contact effectors. We ensure the existence of a quasi-static COM trajectory by restricting our plan to quasi-flat contacts.
For planning 10 steps with less than 10 potential contact surfaces for each phase, our approach is 50 to 100 times faster that its MI counterpart, which suggests potential applications for online contact re-planning.
The method is demonstrated in simulation with the humanoid robots HRP-2 and Talos over various scenarios.

\end{abstract}


\section{Introduction}

Motion planning of legged robots in arbitrary contexts is still an open problem.
Specifically, we are concerned with planning contact locations that allow the robot to move towards its target without falling.


Since the problem is nonlinear, local approaches often lead to dead-ends. Nonlinear optimization formulations are able to provide impressive results at convergence~\cite{Mordatch:2012:DCB:2185520.2185539,Winkler2018}, but the solvers are prone to fall in local minima. Sampling-based approaches~\cite{Hauser06usingmotion,escande:ras:2013}, including our own work~\cite{tonneau-TRO18,carpentier2018multicontact}, as well as A*-based approaches~\cite{4813868,8593694,griffin_arxiv}, have demonstrated significant successes on real robots for specific sets of scenarios.
An appealing approach to tackle the contact planning problem in the most general way is to use mixed-integer (MI) programming. Deits et al. have demonstrated the potential for such approach with a purely kinematic formulation~\cite{DBLP:conf/humanoids/DeitsT14}. Ponton et al. then extended MI to account for the centroidal dynamics of the robots~\cite{ponton:humanoids2016,aceitunocabezas:hal-01674935}. 

MI can theoretically address entirely the \textit{multi-contact planning} problem, extending 
locomotion to non-gaited behaviors, involving the use of hands and other effectors. A practical limitation of MI programming is the computation time, resulting from the branch-and-bound techniques to handle the combinatorial aspect of the problem (here the issue of selecting contact surfaces), especially when the inner problem is hard (such as when nonlinear dynamics have to be handled). Several hundreds of milliseconds are required to plan one or two steps, while ideally we would like to plan the next contact more reactively.

The objective of this paper is to mitigate the issue of computation time for MI, while preserving its potential to tackle contact planning in a general manner.
To achieve this, we propose a convex relaxation of the problem with an L1-norm minimization formulation, which has the desirable property of often converging to sparse solutions~\cite{bach:sparsity2011,Boyd:2004:CO:993483,Skouras:2013}. The sparse solution can then be exploited to decide which surfaces to select.

\begin{figure}
\centering
  \begin{overpic}[width=1\linewidth]{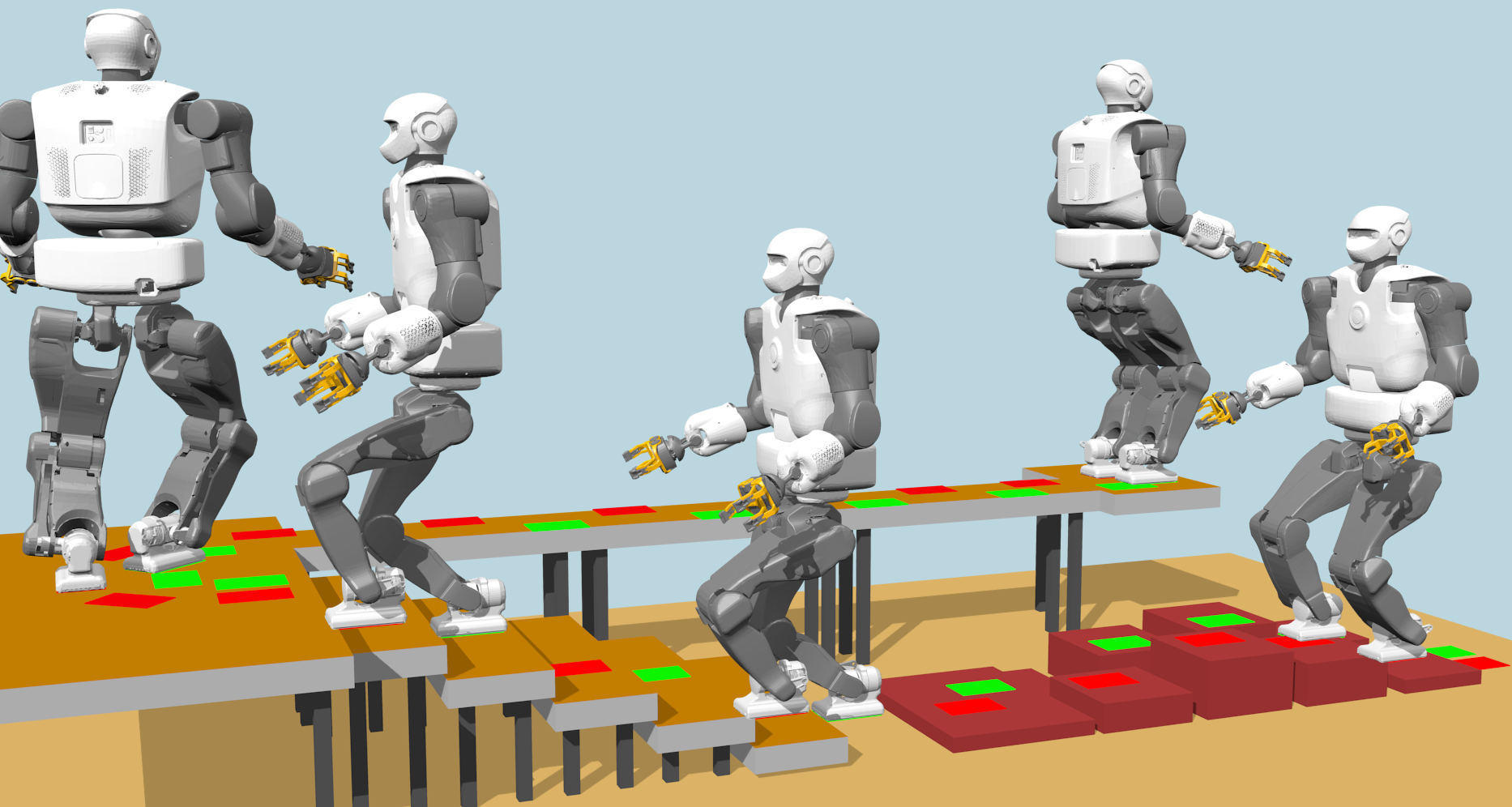}
	\end{overpic}
\caption{A contact sequence computed by our approach, with keyframes of the Talos robot executing the plan.}
		   \label{fig:teaser}
\end{figure}


This reformulation, which we call SL1M (for Sparse L1-norm Minimization) is faster than the MI approach for planning 10 steps with less than 10 potential contact surfaces for each phase, up to 100 times in the most favorable cases. SL1M comes with two drawbacks. First, optimality is only approximated, as the computed solution is a compromise between the imperative of finding a sparse solution and the considered objective function. Second, in some cases involving large numbers of contacts and potential contact surfaces (more than 15 steps and more than 10 potential contact surfaces), the solver can fall in unsatisfying minima (i.e. not converging to entirely sparse solutions), though we believe those minima can be escaped by combining the contact planner with high-level approaches~\cite{Fernbach:iros17}.

\subsection{Contributions}
Our main contribution is the reformulation of the MI contact planning program into a 
convex feasibility linear program (Section~\ref{sec:relax}), for which we provide an analysis of the benefits and drawbacks with respect to MI (Section~\ref{sec:results}).

To guarantee that the contact plans are feasible, SL1M continuously handles the dynamics of the robot using a quasi-static constraint that applies on quasi-flat surfaces (for which the friction cone contains the gravity). This is achieved by extending the 2-PAC approach~\cite{Tonneau:2018:TAC:3278329.3213773} to handle variable foot positions (Section~\ref{sec:reach}), while we leave for future work the extension of the formulation to fully dynamic cases such as~\cite{ponton:humanoids2016}, which is not the focus of this paper.

Lastly, we combine the contact planner with a guide-path planner to automatically initialize the contact planning problem with relevant contact surfaces (Section~\ref{sec:quali}).

\section{Rationale}


Given the initial and final sets of admissible contact postures, $\mathcal{I}$ and $\mathcal{G}$, SL1M computes a sequence of contact locations connecting $\mathcal{I}$ to $\mathcal{G}$ such that a feasible motion exists between each consecutive contact posture. The following assumptions are made:

\begin{itemize}
\item Dynamics constraints are verified by exhibiting a ``quasi-static'' trajectory for the center of mass (COM). 
\item The planner is limited to ``quasi-flat'' contact surfaces (for which the friction cone contains the gravity). This limitation is due to our dynamics approximation and not to the l1-norm approximation.
\item The kinematics constraints on the COM are approximated as linear inequalities attached to the frame of each effector~\cite{7363423,Tonneau:2018:TAC:3278329.3213773}. Similarly the relative positions of the effectors is linearly constrained.
\item The yaw orientations of the contacts at each step is a given, as opposed to~\cite{DBLP:conf/humanoids/DeitsT14}. 
Not handling orientation is a design choice, motivated by the objective to combine the contact planner with a higher-level method (detailed in Section~\ref{sec:quali}). 
\item As for~\cite{DBLP:conf/humanoids/DeitsT14} the planner requires as input a set of convex potential contact surfaces (e.g., they may be given by a sampling-based guide-path planner).
\item The user must provide an initial guess of the order followed by the effectors to create contacts (in our experiments we assume a cyclic gait pattern, which is not limiting for biped robots).
\end{itemize}


\section{Definitions and notations}

\subsection{Important note on the mathematical formulation}
To simplify the equations, we present the problem for the case of gaited bipedal walking. Section~\ref{sec:full_pb_quad} discusses the generalization to acyclic locomotion.


\subsection{Problem variables}
\paragraph*{Contact phases} The motion of the robot is decomposed into a discrete sequence of $n_{ph}$ \textbf{contact phases}. The contact phase $k \in [1, n_{ph}]$ is associated with one or several effectors in contact. Exactly one contact is broken and relocated from phase $k$ to $k+1$. For bipedal walking each contact phase is a double support phase (both feet in contact).\\

\paragraph*{End-effectors} Any effector is identified by an index $j \in [1, n_{eff}]$, where $n_{eff}$ is the number of effectors. The position of the effector moving (swing foot) between phases $k-1$ and $k$ is noted $\mathbf{p}^k = [p_x^k, p_y^k, p_z^k] \in \mathbb{R}^3$. The position of the non-moving effector (support foot) is thus $\mathbf{p}^{k-1}$. $\mathbf{p}^{0}$ is the position of the support effector for the first phase.  \\ 

\paragraph*{Convex hulls} The convex hull of all the effectors in contact at phase $k$ is noted $\mathbf{conv}^k$.\\

\paragraph*{Center Of Mass (COM)} The  COM of the robot $\cv \in \mathbb{R}^3$ moves continuously with a negligible acceleration, under a quasi-static assumption. The quasi-static assumption is only used to provide a certificate of feasibility, such that the resulting motions do not follow this constraint. \\


\paragraph*{Contact surfaces} The possible contact locations are convex surfaces $\mathcal{S}_i$, $1 \leq i \leq n_{surf}$. A surface $\mathcal{S}_i$ is defined by  $n_i + 1$ planes: the plane aligned with the surface defines an equality constraint, while the $n_i$ planes that bound the surface define as many inequality constraints:
\begin{equation*}
\mathcal{S}_i: \{ \vx \in \mathbb{R}^{3},  \vc{S}_i \vx \leq \vc{s}_i, \vc{d}_i^T \vx = e_i  \}
\end{equation*}

\paragraph*{Slack variables} Our formulation uses two kinds of slack variables. Positive variables are denoted  $\boldsymbol\alpha^k = [\alpha_1^k, \dots, \alpha_{n}^k]^T \in \mathbb{R}^{n+}$ and the others $\boldsymbol\beta^k = [\beta_1^k, \dots, \beta_{n}^k]^T \in \mathbb{R}^n$, with $k$ the related contact phase. \\

\paragraph*{Cardinality} The cardinality operator $\mathbf{card}(\boldsymbol\alpha)$ gives the number of entries of $\boldsymbol\alpha$ that are strictly greater than 0.


\section{Contact surface selection constraints}
\label{sec:relax}
When in contact, the position $\mathbf{p}$ of a foot  must belong to a candidate contact surface  $\mathcal{S}_i$.
In the following we detail the convex relaxation of this non-convex constraint.

\subsection{Logical OR as a ``Minimal number of violations'' problem}


We look for a point  $\vx$ belonging to exactly one surface of a set $\mathcal{\mathbf{S}} = \bigcup_{i=1}^{n_{surf}} \mathcal{S}_i$. We assume that the surfaces  $\mathcal{S}_i$ do not intersect. The problem is written:
\begin{equation}\label{eq:or_optim}
\begin{aligned}
\mathbf{find} \quad & \vx, & \\
\st \quad & \vx \in  \mathcal{S}_1 & \lor & &  \vx \in  \mathcal{S}_2 & \dots & \lor & &  \vx \in  \mathcal{S}_{n_{surf}}
\end{aligned}
\end{equation}

With $\lor$ the logical  \textit{or} operator. 
Solving problem (\ref{eq:or_optim}) is equivalent to finding $\vx$ such that as many constraints $\vx \in  \mathcal{S}_i$ as possible are satisfied, which corresponds to $1$ constraint. This optimum is only reached if $\vx$ effectively belongs to one of the surfaces. This is in turn equivalent to finding $\vx$ such that the \textit{minimum number of violations} occurs, which is $n_{surf} - 1$. This can be written as a cardinality minimization problem:

\begin{equation}\label{eq:cardinality}
\begin{aligned}
\mathbf{find} \quad & \vx, \boldsymbol\alpha, \boldsymbol\beta& \\
\mathbf{min}  \quad & \mathbf{card}(\boldsymbol\alpha) & \\
\st \quad & \mathbf{S}_i \vx \leq \mathbf{s}_i + \mathbf{1} \alpha_i & \forall i \\
    \quad & \vc{d}_i^T  \vx = e_i + \beta_i &  \forall i \\
\quad & -\alpha_i \leq \beta_i \leq \alpha_i & \forall i
\end{aligned}
\end{equation}
where $\mathbf{1}$ is a vector of appropriate size filled with 1.
Let us explain why \eqref{eq:cardinality} and \eqref{eq:or_optim} are equivalent.
Assume a given $i$ such that $\bold p \notin \mathcal{S}_i$.
Then, either $\alpha_i \neq 0$ or $\beta_i \neq 0$; moreover, the last constraint of \eqref{eq:cardinality} (equivalent to $|\beta_i| \leq \alpha_i$) implies that if $\beta_i\neq0$, then $\alpha_i\neq0$. 
We conclude that if $\bold p \notin \mathcal{S}_i$ then $\alpha_i \neq 0$.
If instead $\bold p \in \mathcal{S}_i$, then $\beta_i=0$ and $\alpha_i$ is going to be set to zero by the solver because this decreases the objective function.
The function $\mathbf{card}(\boldsymbol\alpha)$ thus counts the number of surfaces on which $\bold p$ does not lie.
Therefore, the optimal value of $\mathbf{card}(\boldsymbol\alpha)$, which is $n_{surf}-1$, is obtained by choosing $\bold p$ to lie on one of the surfaces, which is exactly what~\eqref{eq:or_optim} does.


Rather than addressing this problem with a branch and bound approach (as in MI programming), we can use a convex relaxation of the problem, approximating $\mathbf{card}(\boldsymbol\alpha)$ with the L1-norm $||\boldsymbol\alpha||_1$.
Moreover, since the elements of $\boldsymbol\alpha$ are nonnegative, we have $||\boldsymbol\alpha||_1 = \bold 1^\T \boldsymbol \alpha $.
This approximation of~\eqref{eq:cardinality} is thus a Linear Program:
\begin{equation}\label{eq:card_l1}
\begin{aligned}
\mathbf{min}  \quad & \bold 1^\T \boldsymbol \alpha  & \\
\st \quad & \mathbf{S}_i \vx \leq \mathbf{s}_i + \mathbf{1} \alpha_i & \forall i \\
    \quad & \vc{d}_i^T \vx = e_i + \beta_i &  \forall i \\
\quad & -\alpha_i \leq \beta_i \leq \alpha_i & \forall i
\end{aligned}
\end{equation}
Because L1-minimization leads to sparse solutions, at the optimum we can hope
that one $\alpha_i$ will be (sufficiently close to) 0. This indicates that the surface $\mathcal{S}_i$ has been
selected.


\section{Reachability constraints}
Two kinds of additional constraints are considered in the model.
First, the position of the COM is constrained with respect to the contact points, following the 2-PAC approach~\cite{Tonneau:2018:TAC:3278329.3213773}. This allows to continuously guarantee feasibility of the COM trajectory while only considering
two discrete COM positions at each contact phase. We recall the method for completeness and extend it to handle variable foot translations under the quasi-flat constraint.
Then, the position of each effector is constrained with respect to the other effector in contact.

\subsection{Center Of Mass constraints}

To guarantee that the equilibrium and balance constraints are continuously satisfied for a contact phase $k$, we will use the 2-PAC formulation. 
In the following we recall that we only need to choose 2 COM positions for each phase, namely $\mathbf{c}^{k,0}$ and $\mathbf{c}^{k,1}$, to guarantee continuous feasibility.

\subsubsection{Equilibrium constraints} \label{ssec:equi_con}

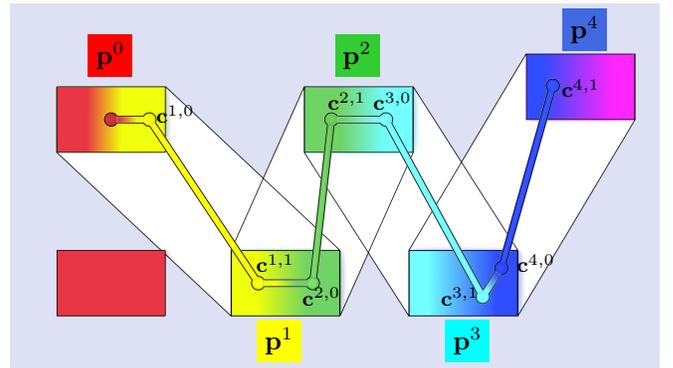
\begin{figure}[b!]
\centering
  \begin{overpic}[width=1\linewidth]{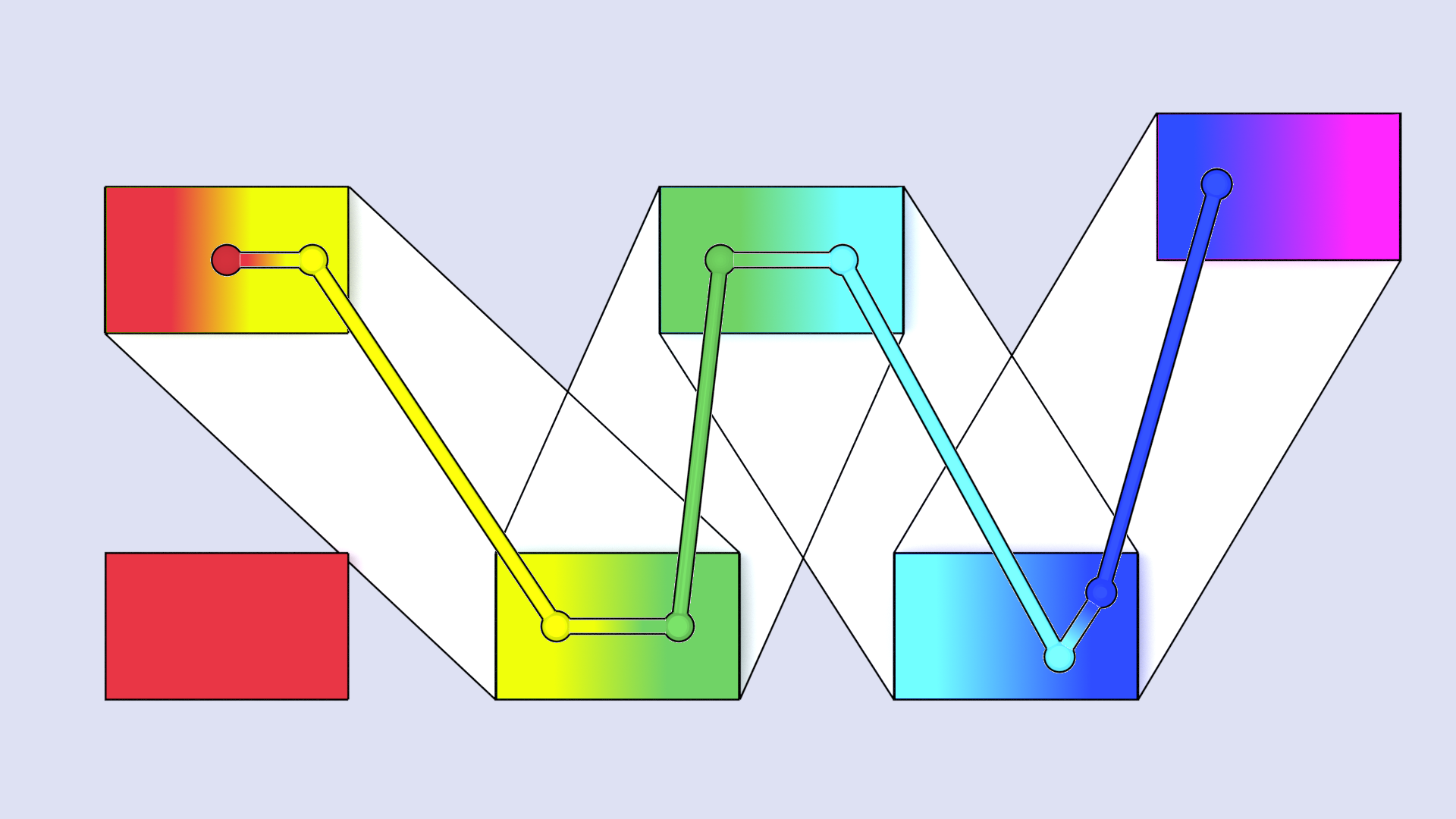}
		\put (12,47) {\colorbox{Red}{$\vx^{0}$}} 
		\put (38,3) {\colorbox{Yellow}{$\vx^{1}$}} 
		\put (50,47) {\colorbox{LimeGreen}{$\vx^{2}$}} 
		\put (67,3) {\colorbox{Cyan}{$\vx^{3}$}} 
		\put (85,51) {\colorbox{RoyalBlue}{$\vx^{4}$}} 
		\put (22.5,37.9) {{\footnotesize$\cv^{1,0}$}} 
		\put (37.9,14.9) {{\footnotesize$\cv^{1,1}$}} 
		\put (45,9.8) {{\footnotesize$\cv^{2,0}$}} 
		\put (48.9,39.9) {{\footnotesize$\cv^{2,1}$}} 
		\put (55.9,39.9) {{\footnotesize$\cv^{3,0}$}} 
		\put (66.3,9.8) {{\footnotesize$\cv^{3,1}$}} 
		\put (78,14.9) {{\footnotesize$\cv^{4,0}$}} 
		\put (84.9,41.9) {{\footnotesize$\cv^{4,1}$}} 
	\end{overpic}
\caption{From an initial double support phase (red), 4 double support contact phases are planned. Each contact is used for 2 consecutive phases. The feasible COM quasi-static trajectory is given by a polyline. Color gradients are used when the COM moves inside the support effector during the simple support phase, and plain colors are used to depict the COM trajectory during a double support phase.}
		   \label{fig:com_traj}
\end{figure}

For quasi-flat contact surfaces, a sufficient condition for the COM to allow for static equilibrium is: \mbox{$\cv^k \in \mathbf{conv}^k$}~\cite{Prete2016}. 
For bipedal walking, this boils down to having the COM on top of the support effector.
In this case $\mathbf{c}^{k,0}$ is constrained to lie above the support polygon of $\mathbf{p}^{k-1}$ (i.e. the support foot used in the transition from phase $k-1$ to $k$, 
which was the swing foot for phase $k-1$) at the beginning of phase $k$. We then constrain $\mathbf{c}^{k,1}$ to be above $\mathbf{p}^{k}$ at the end of phase $k$ (Fig.~\ref{fig:com_traj}):
\begin{equation}\label{eq:equi_f}
\begin{aligned}
\mathbf{F}_i^{k-1} (\mathbf{c}^{k,0} - \mathbf{p}^{k-1} ) & \le \mathbf{f}_i^{k-1} + \mathbf{1} \alpha_i^{k-1} \\
\mathbf{F}_i^{k}(\mathbf{c}^{k,1} - \mathbf{p}^{k} )       & \le \mathbf{f}_i^{k}  + \mathbf{1} \alpha_i^k
\end{aligned}
\end{equation}
where $\mathbf{F}_i^{k}$ and $\mathbf{f}_i^{k}$ are the matrix and vector defining the polygonal shape of the foot associated to phase $k$ on surface $\mathcal{S}_i$.
Note that these constraints depend only on the xy coordinates of the COM and the foot positions.

By convexity of the static equilibrium region, the straight line $[{c}^{k,0},{c}^{k,1}]$ continuously satisfies the static equilibrium constraint. 
Similarly, the straight lines $[{c}^{k-1,1}, {c}^{k,0}]$ and  $[{c}^{k,1}, {c}^{k+1,0}]$ are also feasible because the COM stays above the support effector for all the duration of the single support phase (Fig.~\ref{fig:com_traj}).

\subsubsection{Reachability constraints}
\label{sec:reach}
We additionally constrain  $\mathbf{c}^{k,0}$ and $\mathbf{c}^{k,1}$ to guarantee kinematic reachability. 
 We stress that the kinematic constraints are only approximated here, thus the ``guarantees'' that we mention for feasibility are only valid for this simplified representation of the robot.
The COM positions are linearly constrained as follows.
First, for each effector we compute offline a polytope that approximates the reachable COM workspace: a large number of configurations of the robot are randomly sampled,
and those who are collision-free \textbf{and} correspond to a ``quasi-flat'' pair of contacts are kept. For each of those configurations, the COM is expressed in the frame
of a given effector. The convex hull of all the computed COM positions approximates the COM workspace in the effector frame.
For each effector $j$, we thus obtain a 3D polytope $\leftidx{^j}{\mathcal{R}} : \{ \cv \in \mathbb{R}^{3}, \leftidx{^j}{\vc{R}} \cv \leq \leftidx{^j}{\vc{r}} \} $. 

At contact phase $k$, for each contact surface $\mathcal{S}_i$ the orientation of the foot frame is constant\footnote{The yaw is a given of the problem, while the roll and pitch are given by the surface orientation}. 
We note $\mathcal{R}{^k_{i}}$ the rotated polytope associated with contact $\vx^{k}$ at phase $k$, assuming 
it lies on surface $\mathcal{S}_i$. The translation is variable, thus the constraints depend linearly on the effector positions. Both COM positions $\cv^{k,m}, m \in \{0,1\}$ at phase $k$ are thus constrained by the two active contacts $\vx^{k}$ and $\vx^{k-1}$:
\begin{equation}\label{eq:kin_com}
\begin{aligned}
\vc{R}{_i^l} (\cv^{k,m} - \vx^l) \leq \vc{r}{_i^l} + \mathbf{1} \alpha_i^l && \forall i, \forall l \in \{k-1,k\}
\end{aligned}
\end{equation}

Here again, the slack variable $\boldsymbol\alpha$ is used such that only the constraints related to the selected contact surfaces are applied. By convexity of the kinematic constraints, if they are satisfied for $\mathbf{c}^{k,0}$ and $\mathbf{c}^{k,1}$ for all $k$ then they are continuously satisfied.

\subsection{Relative foot position constraints}
Similarly to the case of the COM reachability, we use a sampling-based approach to approximate the reachable workspace of each foot with respect to the others.
For effector $j$, we obtain a polytope $\leftidx{^j}{\mathcal{Q}} : \{ \vx \in \mathbb{R}^{3}, \leftidx{^j}{\vc{Q}} \vx \leq \leftidx{^j}{\vc{q}} \} $ that constrains the other effector. If $j$ is the moving effector at phase $k$ on surface $\mathcal{S}_i$, we abusively write $\mathcal{Q}^k_i = \leftidx{^j}{\mathcal{Q}}_i$ for clarity.
We then apply the same reasoning as for the COM to obtain the following constraints at each phase:
\begin{equation}\label{eq:kin_rel}
\begin{aligned}
\vc{Q}{^{k-1}_i} (\vx^k - \vx^{k-1}) \leq \vc{q}{^{k-1}_i} + \mathbf{1} \alpha_i^{k-1} &&  \forall i
\end{aligned}
\end{equation}
\subsection{An important simplification}
If the contact surfaces for the effectors are punctual (as for most quadruped robots), or if the candidate surfaces share the same orientation, then the polytope constraints $\mathcal{R}{^k_{i}}$ (respectively $\mathcal{Q}{^k_i}$ and $\mathcal{F}_i$) are the same for all $i$. In this case we do not need to use the selection variable $\boldsymbol\alpha$ in equations~\ref{eq:equi_f},~\ref{eq:kin_com} and~\ref{eq:kin_rel}.

\section{The complete feasibility problem}
\label{sec:full_pb}
Combining all the constraints, we can now write the complete feasibility problem that defines SL1M. For the sake of simplicity, and without loss of generality,
we assume that all contact surfaces are potential candidates for all phases.    
\begin{equation}\label{eq:complete_problem}
\begin{aligned}
\mathbf{find} \quad & \mathbf{u} = [\vc{p}^1, \dots , \vc{p}^{n_{ph}}, \boldsymbol\alpha^1 , \dots , \boldsymbol\alpha^{n_{ph}},  \boldsymbol\beta^1 , \dots , \boldsymbol\beta^{n_{ph}}, \\ 
\quad & \vc{c}^{1,0}, \dots , \vc{c}^{n_{ph},0} , \vc{c}^{1,1}, \dots , \vc{c}^{n_{ph},1} ]\\ 
\mathbf{min} \quad & \sum_{k = 1}^{n_{ph}}  \vc{1}^T\boldsymbol\alpha^k + \gamma l(\mathbf{u})\\ 
\st \quad &\{\vc{p}^1, \cv^{1,0}\} \in \mathcal{I} \\ 
\quad  & \{\vc{p}^{n_{ph} - 1}, \vc{p}^{n_{ph}}, \cv^{n_{ph},1}\} \in \mathcal{G} \\ 
\quad & \forall \, k,  i: \\
        \quad & \quad \quad \mathbf{S}_i \vx^k \leq \mathbf{s}_i + \mathbf{1} \alpha_i^k & \\
        \quad & \quad \quad \vc{d}_i^T \vx^k = e_i + \beta_i^k &  \\
        \quad & \quad \quad -\alpha_i^k \leq \beta_i^k \leq \alpha_i^k  \\        
        \quad & \quad \quad \mathbf{F}_i^{k-1} (\mathbf{c}^{k,0} - \mathbf{p}^{k-1} ) \le \mathbf{f}^i_{k-1} + \mathbf{1} \alpha_i^{k-1} \\
        \quad & \quad \quad \mathbf{F}_i^{k}(\mathbf{c}^{k,1} - \mathbf{p}^{k} )      \le \mathbf{f}_i^{k}  + \mathbf{1} \alpha_i^k \\
        \quad & \quad \quad {\vc{Q}}{^k_{i}} (\vx^k - \vx^{k-1}) \leq {\vc{q}}{^k_{i}} + \mathbf{1} \alpha_i^k \\
        \quad & \quad \quad  \forall \, l \in \{k-1,k\}, \forall \, m \in \{0,1\} : \\
        \quad & \quad \quad \quad  \quad \vc{R}{_i^l} (\cv^{k,m} - \vx^l) \leq \vc{r}{_i^l} + \mathbf{1} \alpha_i^l \\
\end{aligned}
\end{equation}
where $l$ is an optional quadratic or linear objective function and $\gamma$ is a small weighing value.

$ \mathcal{I}$ and $ \mathcal{G}$ define initial and goal state conditions. The initial/goal values for contact locations and COM can be given exactly, or loosely specified.  For instance the last contact can be constrained to lie on a given contact surface, which is convenient for specifying the goal state in general.


\section{Extension to Acyclic Multi-Contact Locomotion}
\label{sec:full_pb_quad}
So far we have presented an approach to plan contacts for bipedal locomotion on quasi-flat terrains.
However, we believe that this approach can be extended to multi-contact locomotion, as long as contact surfaces remain quasi-flat.

To do this, the main change needed regards the equilibrium constraints (Section~\ref{ssec:equi_con}): \mbox{$\cv^k \in \mathbf{conv}^k$}.
If we assume point-like contacts, these constraints can be written as:
\begin{equation}\label{eq:equi_multi_contact}
\begin{aligned}
\mathbf{c}^k = \sum_j w_j^k \vx^k_j, \qquad \sum_j w_j^k = 1, \qquad w_j^k \ge 0 \quad \forall j,
\end{aligned}
\end{equation}
where the $\vx^k_j$'s are the positions of the effectors in contact at phase $k$, and $\mathbf{w}^k$ is a unit weighting vector.
Since the contact positions are variable, \eqref{eq:equi_multi_contact} is bilinear, hence non-convex.
To maintain the formulation convex, we could choose a conservative approach.
For each phase we can fix $\mathbf{w}^k$, such that the COM xy coordinates are directly determined from the contact positions.
A similar approach could be used for finite-size contact surfaces: we can express the COM as a convex combination of virtual contact points $\bar{\mathbf{p}}_j^k$, which must lie inside the associated effector surfaces with position $\vx^k_j$:
\begin{equation}\label{eq:equi_com}
\begin{aligned}
\mathbf{c}^k &= \sum_j w_j^k \bar \vx^k_j \\
\mathbf{F}_i^{k}(\bar \vx^k_j  - \mathbf{p}^{k} )       & \le \mathbf{f}_i^{k}  + \mathbf{1} \alpha_i^k
\end{aligned}
\end{equation}

Apart from this new formulation of the equilibrium constraints, problem~\eqref{eq:complete_problem} could be used almost as it is for multi-contact problems---even though we plan to test this extension in future work.

\section{Fixing the sparsity}

\subsection{Near sparse convergence}
\label{sec:fix_spar}
Once problem~(\ref{eq:complete_problem}) is solved, in the ideal case, for each phase $k$ one and only one $\alpha_{i}^k$ value is equal or sufficiently close to 0.
We can then \textit{fix the sparsity} by assigning to each phase its selected surface and solve problem~(\ref{eq:complete_problem}) without the slacks $\boldsymbol\alpha$ and $\boldsymbol\beta$ again to obtain an exact solution. 

It might be the case that for some phases no $\alpha_i^k$ is close enough to 0. This indicates either that the problem is unfeasible or that the convex relaxation leads to a non-sparse optimum. If the number of concerned phases is low, the non-sparse minimum might be close enough to a sparse optimum. We then solve a small combinatorial problem. We test all the possible combinations of surface selection for the unresolved phases by assigning one and only one single contact surface to each unresolved phase, and we solve again problem~(\ref{eq:complete_problem}). We can either test all combinations or stop at the first feasible solution, in which case
the resolution remains computationally efficient.


In our tests we noticed that, if the problem was feasible, choosing the surfaces associated to the lowest 
$\alpha_i^k$ led to the desired sparse solution in the majority of cases.

\subsection{Handling non-sparse optimum}
If too many phases are unresolved (i.e. no $\alpha_i^k \approx 0$) the non-sparse optimum might be too far from a sparse optimum. 

This happens with a brute force formulation of our main test scenario (Fig.~\ref{fig:teaser}), where the  start and goal configurations are spatially close, but a long path is required to connect them. This is a classical case where nonlinear solvers would fail by trying to directly connect the states.

This issue can be mitigated by initializing the solver with a relevant initial guess. Fortunately an intuitive solution can be built either manually or automatically,
by limiting the set of contact surfaces available at each phase.
If we separate the scene into 2 distinct sets of potential contact surfaces (``before'' and ``after'' the top of the staircase), the solver is able to converge to a valid solution.
 
Proper and automatic initialization is thus essential for large problems, and discussed in Section~\ref{sec:quali}.

\section{Results}
\label{sec:results}
SL1M has been successfully tested in simulation for two humanoid robots with different kinematic constraints, namely HRP-2 and Talos \cite{stasse:talos}.
In this section we also present a quantitative comparison of the performances of our approach with respect to a mixed integer implementation.



\subsection{Implementation details}
Problem~(\ref{eq:complete_problem}) has been implemented in python, using the sparse, open source linear solver GLPK~\cite{makhorin2008glpk} \footnote{Once the sparsity is fixed, we use Quadprog with a quadratic cost that favors contact positions at the center of the contact surfaces}.
For comparison the problem has also been implemented in its MI form using gurobi~\cite{gurobi}, interfaced with cvxpy~\cite{cvxpy}. 
Tests were run on a laptop with an i7 processor on ubuntu 16.04. Our code will be open sourced upon acceptance of the paper.


\subsection{Quantitative analysis and comparison with MI}
To determine empirically the capabilities of SL1M and compare it with the MI approach, we designed two experiments and computed the ratio
between the obtained computation times for each formulation. 
The computation times for SL1M include the extra steps required to fix the sparsity described in Section~\ref{sec:fix_spar}.
The presented times are averaged over 10 runs.



\begin{figure}[]
\centering
\begin{subfloat}
  \centering
 \begin{overpic}[width=1\linewidth]{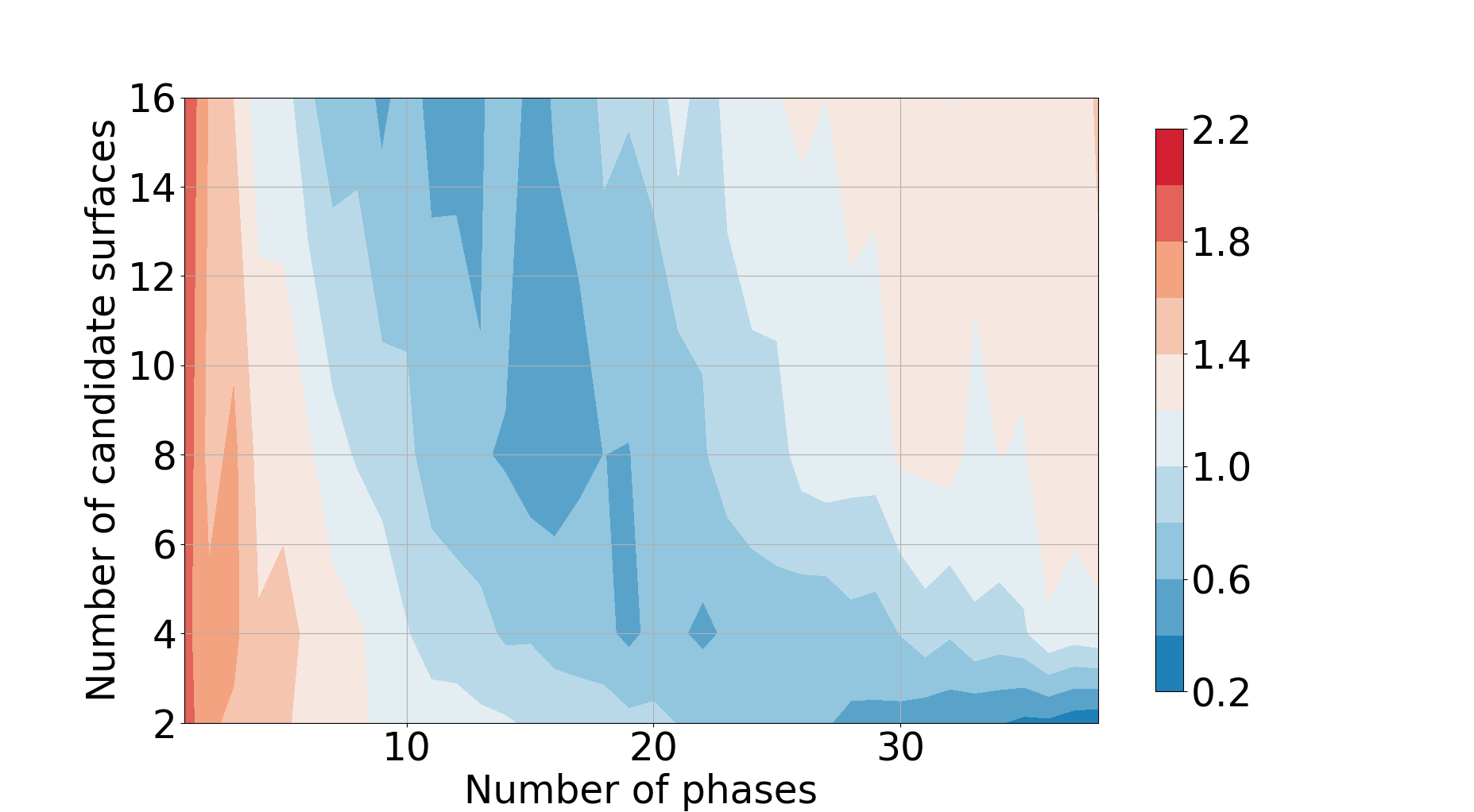}
\end{overpic}
\end{subfloat}
\begin{subfloat}
  \centering
 \begin{overpic}[width=1\linewidth]{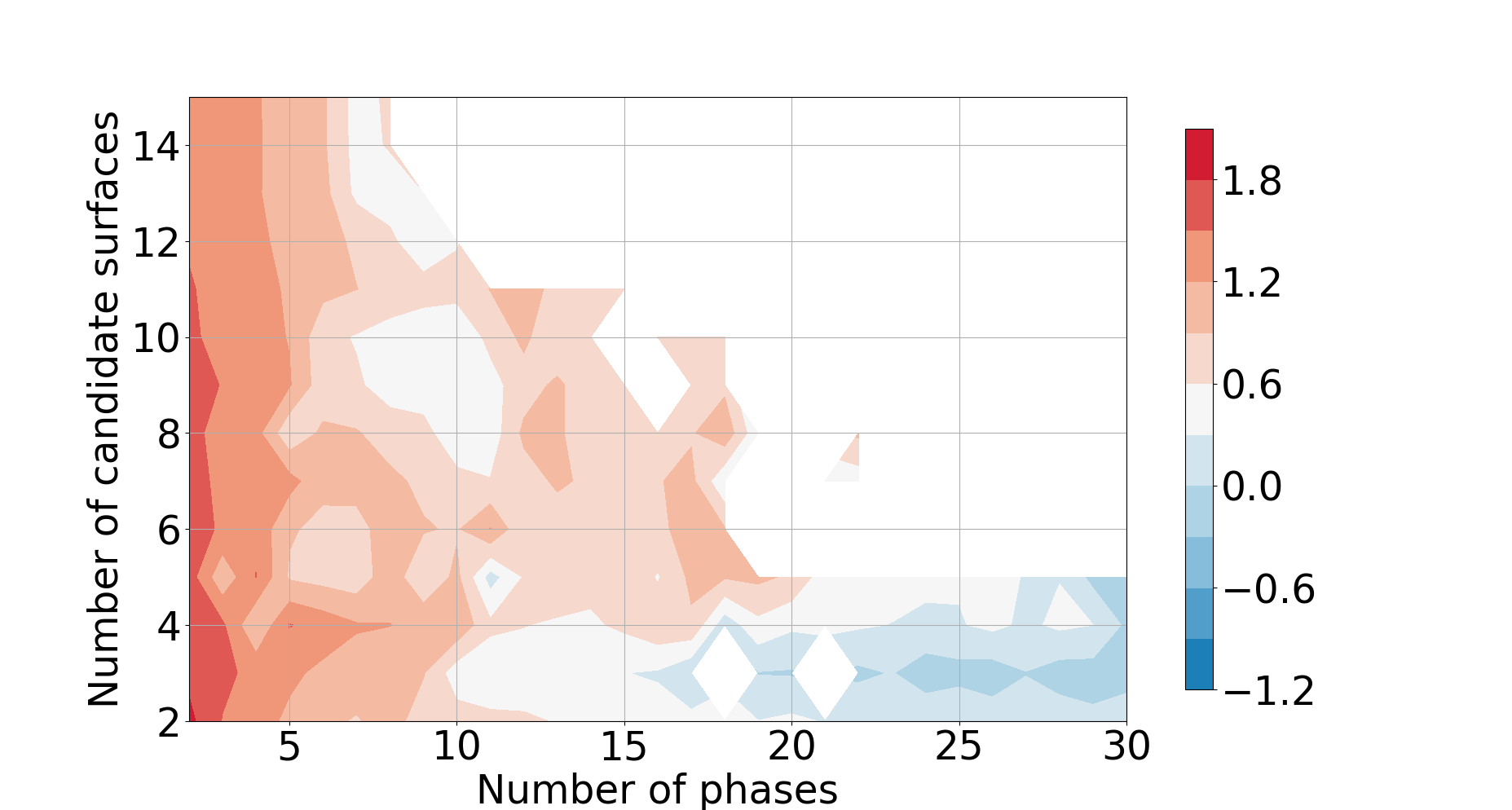}
\end{overpic}
\end{subfloat}%
\caption{Ratio of computation times obtained for SL1M over the times for the MI formulation for the toy planar scenario (up) and the LAAS experiment room (down) in logarithmic 10 scale.}
\label{fig:bench}
\end{figure}

\subsubsection*{Toy problem} We consider an environment composed of a single flat surface, on which the robot is required to walk between 2 and 38 steps.
Variations of this scenario involve iteratively splitting the surface to artificially create more contact candidates.

The results are presented by the color map in Fig.~\ref{fig:bench} - top. 
The computation times range from 50 ms to about 17 s for MI, and from 0.47 ms to 585 ms for SL1M.
In this scenario, SL1M is always faster, with a speed up factor between 1.5 and 100.
The most favorable setups are the smaller problems.

We conclude that, for this toy problem, SL1M outperforms MI in finding a feasible contact sequence, without any significant drawback.

\subsubsection*{Challenging scenario}
We now consider the LAAS experiment room shown in Fig.~\ref{fig:teaser}. Again, we change the size of the problem by changing
the number of steps and possible contact surfaces. To achieve this, we start from a known feasible solution of the problem connecting the first and last poses of Fig.~\ref{fig:teaser}. We obtain 30 contact phases with exactly one contact candidate for each phase. To increase the number of surface candidates for a phase $k$, we add those of the surrounding phases $k-1$ and $k+1$, and iterate as required. The problem is thus always feasible. To reduce the number of steps, we simply cut the last phases.

The results are presented by the color map in Fig.~\ref{fig:bench} - bottom. For this scenario, when the L1-norm did not converge to a sparse solution, we decided to mark the scenario as unsolved if the combinatorial involved more than 4000 possible cases, otherwise we let the combinatorial run (with the possibilities sorted according to the $\alpha_i^k$ values) until a feasible solution was found. The upper right part of the plot is thus left blank, indicating a ``failure'' of the L1-norm. On the other hand MI was always successful.
In this case the computation times range from 43 ms to 15 s for MI, and from less than 1 ms to 3.5 s for SL1M.
Fig.~\ref{fig:bench} shows some cases where the MI is faster (up to 10 times) as the number of phases increase. The drawback of SL1M is also highlighted 
by the failures registered for large problems. However we still observe that SL1M outperforms significantly MI for small and medium problems (up to 40 times faster).

These results suggest that SL1M is especially suited for scenarios involving few potential contact surfaces at each phase, while additional work is required to analyze in depth the failures that occur for larger problems, and propose a mean to address them, for instance by considering stricter norms to enforce the sparsity~\cite{Skouras:2013}.

\subsection{Qualitative validation of the contact sequence}
\label{sec:quali}
To validate SL1M qualitatively, the generated contact sequences are given as input to an open-source whole-body motion generator\footnote{https://github.com/loco-3d/multicontact-locomotion-planning} presented in~\cite{carpentier2017multi,Fernbach:ccroc}. 
For lack of space we cannot describe this generator in detail here. 

\begin{figure}[]
\centering
\begin{subfloat}
  \centering
 \begin{overpic}[width=.7\linewidth]{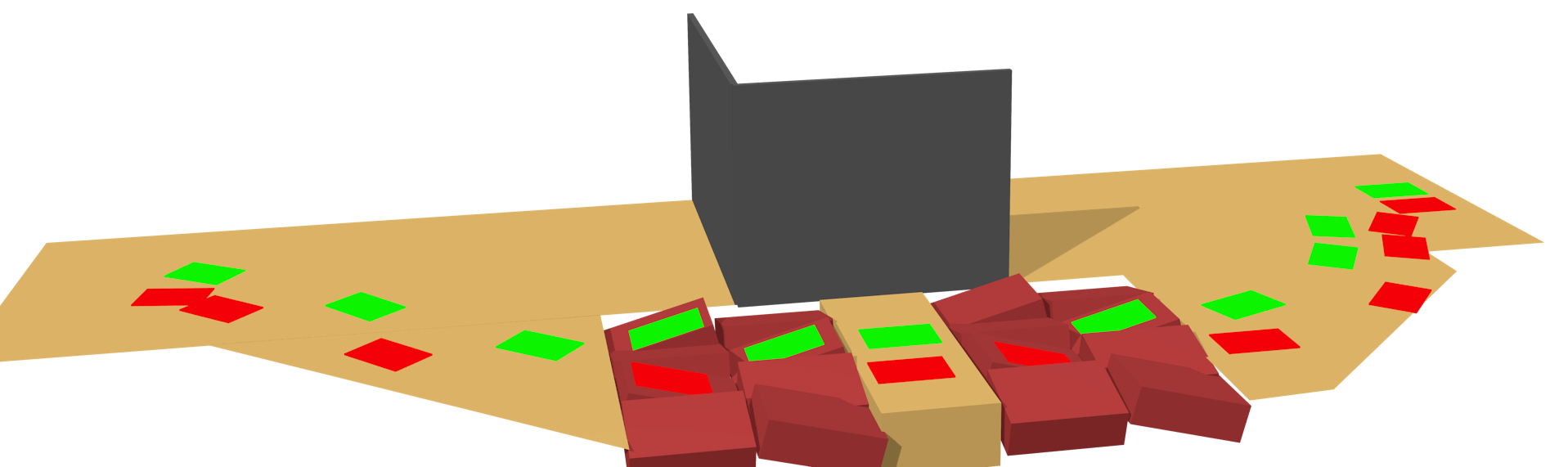}
\end{overpic}
\end{subfloat}
\begin{subfloat}
  \centering
 \begin{overpic}[width=.29\linewidth]{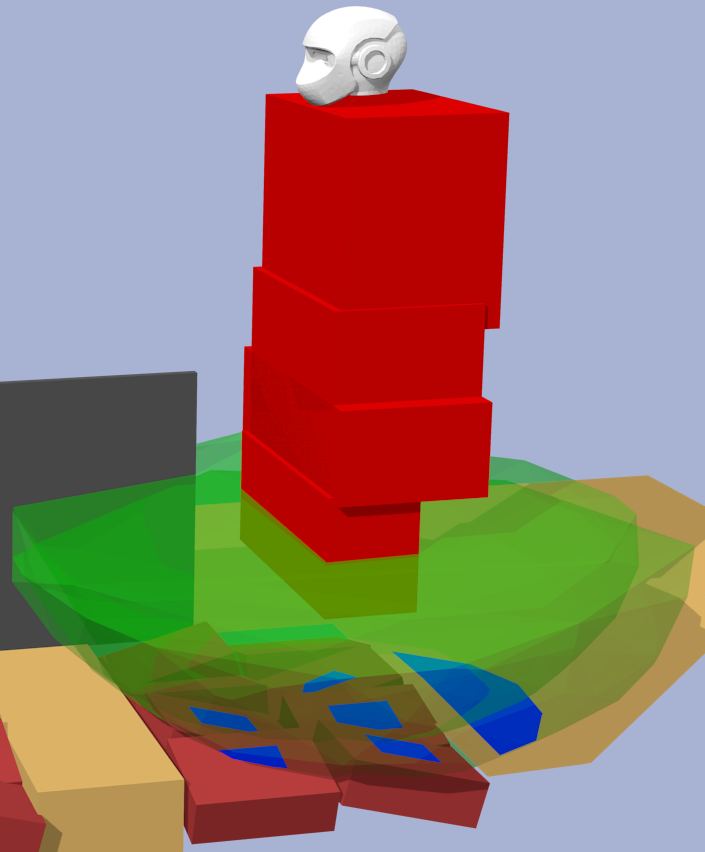}
\end{overpic}
\end{subfloat}%
\caption{Left: Contact sequences on uneven surfaces for Talos. Right: For the root location, the approximated reachable workspace of the feet (green) gives a set of potential contact surfaces (blue). The right shape approximates the collision avoidance constraint, similarly to~\cite{DBLP:conf/humanoids/DeitsT14}.}
		   \label{fig:uneven}
\end{figure}

In our demonstrations, rather than manually providing the potential contact surfaces and orientations, we initialize~(\ref{eq:complete_problem}) with a sampling-based guide path planner~\cite{Fernbach:iros17}. Given a 3D mesh as well as start and goal configurations for the root of the robot,
the planner computes a 6D collision-free trajectory for this root. The planner is able to return, for each position of the root along the path, a set of potential contact surfaces for each effector (Fig.~\ref{fig:uneven} - right). Because the contact surfaces correspond to collision-free positions of the root, the output of the contact planner is more likely to avoid collisions (Fig.~\ref{fig:uneven}). The guide path is thus discretized into $n_{ph}$ contact phases, from which we extract the potential contacts for each phase. The root yaw orientation at each phase $k$ becomes the orientation constraint for the moving foot $\vx^k$.



\begin{figure}[]
\centering
\begin{subfloat}
  \centering
 \begin{overpic}[width=.8\linewidth]{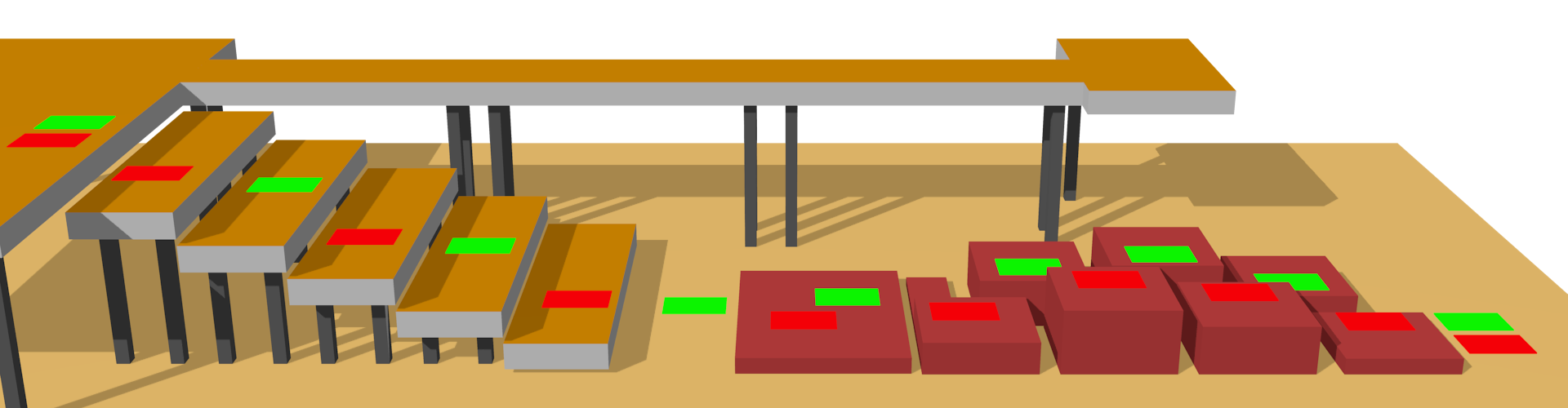}
\end{overpic}
\end{subfloat}
\begin{subfloat}
  \centering
 \begin{overpic}[width=.8\linewidth]{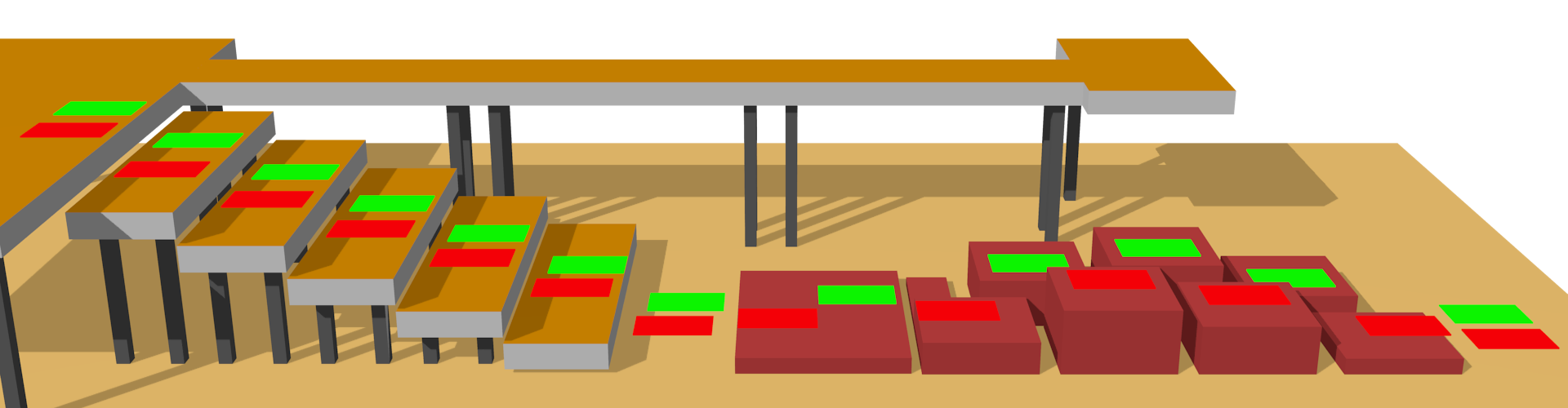}
\end{overpic}
\end{subfloat}%
\caption{Example of contact sequences in the LAAS experimental room for Talos (top) and HRP-2 (bottom). Interestingly the kinematic constraints of each robot lead to different strategies.}
\label{fig:talos_hrp2}
\end{figure}

The companion video demonstrates that contact plans such as the ones displayed in Fig.~\ref{fig:teaser},~\ref{fig:uneven} and~\ref{fig:talos_hrp2} can be extended to dynamically consistent and collision-free motions. The motions with Talos have been validated on the simulator Gazebo, using a torque controller provided by the robot's manufacturer PAL robotics.


\section{Conclusions and future work}

We present SL1M, a convex relaxation of the mixed integer (MI) programming approach for planning contact sequences for legged robots.
For the issue of finding a feasible contact sequence, L1-norm minimization seems particularly well suited for small and medium size problems, where SL1M outperforms
MI by at least one order of magnitude, at the cost of relegating optimality to a secondary objective. 



We plan to extend the presented approach and its assessment in different ways.
First, our comparison with the MI approach has shown that any approach can outperform the other one depending on the situation.
We want to carry out further tests to understand under which conditions the L1 optimization fails to provide a sparse solution, and explore whether
it could be enforced using stricter norm formulations~\cite{Skouras:2013}.

Another promising direction is the integration of contact planning and guide-path planning, which we have already started to explore in this paper. Using a guide-path planner we can drastically reduce the number of candidate contact surfaces for each contact. This greatly simplifies the contact planning problem, and can make the difference between failure and success. However, it also reduces the exploration space of the contact planner, which may be detrimental. Quantifying the benefits of guide-path planning is subject of ongoing investigation.

While this work has focused on bipedal walking, the presented approach can be applied to any acyclic multi-contact locomotion (as long as contacts are quasi-flat, see Section~\ref{sec:full_pb_quad}). Thus, we plan to test our method with other gaits and with quadruped robots.
In the future we would also like to extend the formulation to integrate fully dynamic constraints in the planning by using a convex formulation
of the problem~\cite{Fernbach:iros18}.

Finally, other robotics problems such as collision-free inverse kinematics~\cite{dai_ik_19} or trajectory planning~\cite{Deits2015EfficientMP, DBLP:journals/corr/abs-1811-10753} can be addressed using Mixed Integer programming. Investigating the use of L1-norm minimization for those problems provides an exciting direction for further research.

\section{Acknowledgements}
The authors would like to thank Joseph Mirabel and Jean-Michel Tonneau for their help, as well as PAL-Robotics for providing a simulator for Talos.

This work is supported by the National Research Foundation (NRF) in South Korea (2017R1A2B3012701) and the H2020 project Memmo (ICT-780l684).

\bibliographystyle{IEEEtran}
\bibliography{IEEEabrv,biblio}

\begin{thebibliography}{10}
\providecommand{\url}[1]{#1}
\csname url@samestyle\endcsname
\providecommand{\newblock}{\relax}
\providecommand{\bibinfo}[2]{#2}
\providecommand{\BIBentrySTDinterwordspacing}{\spaceskip=0pt\relax}
\providecommand{\BIBentryALTinterwordstretchfactor}{4}
\providecommand{\BIBentryALTinterwordspacing}{\spaceskip=\fontdimen2\font plus
\BIBentryALTinterwordstretchfactor\fontdimen3\font minus
  \fontdimen4\font\relax}
\providecommand{\BIBforeignlanguage}[2]{{%
\expandafter\ifx\csname l@#1\endcsname\relax
\typeout{** WARNING: IEEEtran.bst: No hyphenation pattern has been}%
\typeout{** loaded for the language `#1'. Using the pattern for}%
\typeout{** the default language instead.}%
\else
\language=\csname l@#1\endcsname
\fi
#2}}
\providecommand{\BIBdecl}{\relax}
\BIBdecl

\bibitem{Mordatch:2012:DCB:2185520.2185539}
I.~Mordatch, E.~Todorov, and Z.~Popovi\'{c}, ``{Discovery of complex behaviors
  through contact-invariant optimization},'' \emph{ACM Trans. on Graph.},
  vol.~31, no.~4, pp. 43:1--43:8, 2012.

\bibitem{Winkler2018}
\BIBentryALTinterwordspacing
A.~W. Winkler, C.~D. Bellicoso, M.~Hutter, and J.~Buchli, ``{Gait and
  Trajectory Optimization for Legged Systems through Phase-based End-Effector
  Parameterization},'' \emph{IEEE Robotics and Automation Letters}, pp. 1--1,
  2018. [Online]. Available: \url{http://ieeexplore.ieee.org/document/8283570/}
\BIBentrySTDinterwordspacing

\bibitem{Hauser06usingmotion}
K.~Hauser, T.~Bretl, K.~Harada, and J.-C. Latombe, ``Using motion primitives in
  probabilistic sample-based planning for humanoid robots.'' in \emph{WAFR},
  ser. Springer Tracts in Advanced Robot., S.~Akella, N.~M. Amato, W.~H. Huang,
  and B.~Mishra, Eds., vol.~47.\hskip 1em plus 0.5em minus 0.4em\relax
  Springer, 2006.

\bibitem{escande:ras:2013}
\BIBentryALTinterwordspacing
A.~Escande, A.~Kheddar, and S.~Miossec, ``Planning contact points for humanoid
  robots,'' \emph{Robotics and Autonomous Systems}, vol.~61, no.~5, pp. 428 --
  442, 2013. [Online]. Available:
  \url{http://www.sciencedirect.com/science/article/pii/S0921889013000213}
\BIBentrySTDinterwordspacing

\bibitem{tonneau-TRO18}
S.~Tonneau, A.~{Del Prete}, J.~Pettr{\'e}, C.~Park, D.~Manocha, and N.~Mansard,
  ``An efficient acyclic contact planner for multiped robots,'' \emph{IEEE
  Transactions on Robotics}, vol.~34, no.~3, pp. 586--601, June 2018.

\bibitem{carpentier2018multicontact}
J.~Carpentier and N.~Mansard, ``Multicontact locomotion of legged robots,''
  \emph{IEEE Transactions on Robotics}, vol.~34, no.~6, pp. 1441--1460, 2018.

\bibitem{4813868}
J.~{Chestnutt}, K.~{Nishiwaki}, J.~{Kuffner}, and S.~{Kagami}, ``An adaptive
  action model for legged navigation planning,'' in \emph{2007 7th IEEE-RAS
  International Conference on Humanoid Robots}, Nov 2007, pp. 196--202.

\bibitem{8593694}
Y.~{Lin} and D.~{Berenson}, ``Humanoid navigation planning in large
  unstructured environments using traversability - based segmentation,'' in
  \emph{2018 IEEE/RSJ International Conference on Intelligent Robots and
  Systems (IROS)}, Oct 2018, pp. 7375--7382.

\bibitem{griffin_arxiv}
\BIBentryALTinterwordspacing
R.~J. Griffin, G.~Wiedebach, S.~McCrory, S.~Bertrand, I.~Lee, and J.~E. Pratt,
  ``Footstep planning for autonomous walking over rough terrain,'' \emph{CoRR},
  vol. abs/1907.08673, 2019. [Online]. Available:
  \url{http://arxiv.org/abs/1907.08673}
\BIBentrySTDinterwordspacing

\bibitem{DBLP:conf/humanoids/DeitsT14}
R.~Deits and R.~Tedrake, ``Footstep planning on uneven terrain with
  mixed-integer convex optimization,'' in \emph{Humanoid Robots (Humanoids),
  14th IEEE-RAS Int. Conf. on}, Madrid, Spain, 2014.

\bibitem{ponton:humanoids2016}
B.~Ponton, A.~Herzog, S.~Schaal, and L.~Righetti, ``A convex model of humanoid
  momentum dynamics for multi-contact motion generation,'' in \emph{Proceedings
  of the 2016 IEEE-RAS International Conference on Humanoid Robots}, 2016.

\bibitem{aceitunocabezas:hal-01674935}
\BIBentryALTinterwordspacing
B.~Aceituno-Cabezas, C.~Mastalli, H.~Dai, M.~Focchi, A.~Radulescu, D.~Caldwell,
  J.~Cappelletto, J.~C. Grieco, G.~Fern{\'a}ndez-L{\'o}pez, and C.~Semini,
  ``{Simultaneous Contact, Gait and Motion Planning for Robust Multi-Legged
  Locomotion via Mixed-Integer Convex Optimization},'' \emph{{IEEE Robotics and
  Automation Letters}}, pp. 1 -- 1, Dec. 2017. [Online]. Available:
  \url{https://hal.archives-ouvertes.fr/hal-01674935}
\BIBentrySTDinterwordspacing

\bibitem{bach:sparsity2011}
J.~M. F.~Bach, R.~Jenatton and G.~Obozinski, ``Convex optimization with
  sparsity-inducing norms,'' \emph{Optimization for Machine Learning}, 2011.

\bibitem{Boyd:2004:CO:993483}
\BIBentryALTinterwordspacing
S.~Boyd and L.~Vandenberghe, \emph{Convex Optimization}.\hskip 1em plus 0.5em
  minus 0.4em\relax Cambridge University Press, 2004. [Online]. Available:
  \url{https://web.stanford.edu/~boyd/cvxbook/}
\BIBentrySTDinterwordspacing

\bibitem{Skouras:2013}
\BIBentryALTinterwordspacing
M.~Skouras, B.~Thomaszewski, S.~Coros, B.~Bickel, and M.~Gross, ``Computational
  design of actuated deformable characters,'' \emph{ACM Trans. Graph.},
  vol.~32, no.~4, pp. 82:1--82:10, Jul. 2013. [Online]. Available:
  \url{http://doi.acm.org/10.1145/2461912.2461979}
\BIBentrySTDinterwordspacing

\bibitem{Fernbach:iros17}
P.~Fernbach, S.~Tonneau, A.~{Del Prete}, and M.~Ta\"{i}x, ``A kinodynamic
  steering-method for legged multi-contact locomotion,'' in \emph{IEEE/RSJ
  International Conference on Intelligent Robots and Systems (IROS)}, Sept
  2017, pp. 3701--3707.

\bibitem{Tonneau:2018:TAC:3278329.3213773}
\BIBentryALTinterwordspacing
S.~Tonneau, P.~Fernbach, A.~{Del Prete}, J.~Pettr{\'e}, and N.~Mansard, ``2pac:
  Two-point attractors for center of mass trajectories in multi-contact
  scenarios,'' \emph{ACM Trans. Graph.}, vol.~37, no.~5, pp. 176:1--176:14,
  Oct. 2018. [Online]. Available: \url{http://doi.acm.org/10.1145/3213773}
\BIBentrySTDinterwordspacing

\bibitem{7363423}
C.~{Brasseur}, A.~{Sherikov}, C.~{Collette}, D.~{Dimitrov}, and P.~{Wieber},
  ``A robust linear mpc approach to online generation of 3d biped walking
  motion,'' in \emph{2015 IEEE-RAS 15th International Conference on Humanoid
  Robots (Humanoids)}, Nov 2015, pp. 595--601.

\bibitem{Prete2016}
A.~{Del Prete}, S.~Tonneau, and N.~Mansard, ``{Fast Algorithms to Test Robust
  Static Equilibrium for Legged Robots},'' in \emph{2016 IEEE International
  Conference on Robotics and Automation (ICRA)}, Stockholm, Sweden, 2016.

\bibitem{stasse:talos}
\BIBentryALTinterwordspacing
O.~Stasse, T.~Flayols, R.~Budhiraja, K.~Giraud-Esclasse, J.~Carpentier,
  J.~Mirabel, A.~{Del Prete}, P.~Sou{\`e}res, N.~Mansard, F.~Lamiraux, J.-P.
  Laumond, L.~Marchionni, H.~Tome, and F.~Ferro, ``{TALOS: A new humanoid
  research platform targeted for industrial applications},'' in
  \emph{{International Conference on Humanoid Robotics, ICHR, Birmingham
  2017}}, ser. IEEE-RAS 17th International Conference on Humanoid Robotics
  (Humanoids),.\hskip 1em plus 0.5em minus 0.4em\relax Birmingham, United
  Kingdom: {IEEE}, Nov. 2017. [Online]. Available:
  \url{https://hal.archives-ouvertes.fr/hal-01485519}
\BIBentrySTDinterwordspacing

\bibitem{makhorin2008glpk}
A.~Makhorin, ``Glpk (gnu linear programming kit),'' \emph{http://www. gnu.
  org/s/glpk/glpk. html}, 2008.

\bibitem{gurobi}
\BIBentryALTinterwordspacing
L.~Gurobi~Optimization, ``Gurobi optimizer reference manual,'' 2019. [Online].
  Available: \url{http://www.gurobi.com}
\BIBentrySTDinterwordspacing

\bibitem{cvxpy}
S.~Diamond and S.~Boyd, ``{CVXPY}: A {P}ython-embedded modeling language for
  convex optimization,'' \emph{Journal of Machine Learning Research}, vol.~17,
  no.~83, pp. 1--5, 2016.

\bibitem{carpentier2017multi}
J.~Carpentier, A.~Del~Prete, S.~Tonneau, T.~Flayols, F.~Forget, A.~Mifsud,
  K.~Giraud, D.~Atchuthan, P.~Fernbach, R.~Budhiraja \emph{et~al.},
  ``Multi-contact locomotion of legged robots in complex environments--the
  loco3d project,'' 2017.

\bibitem{Fernbach:ccroc}
\BIBentryALTinterwordspacing
P.~Fernbach, S.~Tonneau, O.~Stasse, J.~Carpentier, and M.~Ta{\"i}x, ``{C-CROC:
  Continuous and Convex Resolution of Centroidal dynamic trajectories for
  legged robots in multi-contact scenarios},'' Aug. 2019, working paper or
  preprint. [Online]. Available: \url{https://hal.laas.fr/hal-01894869}
\BIBentrySTDinterwordspacing

\bibitem{Fernbach:iros18}
P.~Fernbach, S.~Tonneau, and M.~Ta\"{i}x, ``Croc: Convex resolution of
  centroidal dynamics trajectories to provide a feasibility criterion for the
  multi contact planning problem,'' in \emph{IEEE/RSJ International Conference
  on Intelligent Robots and Systems (IROS)}, 2018.

\bibitem{dai_ik_19}
\BIBentryALTinterwordspacing
H.~Dai, G.~Izatt, and R.~Tedrake, ``Global inverse kinematics via mixed-integer
  convex optimization,'' \emph{The International Journal of Robotics Research},
  vol.~0, no.~0, p. 0278364919846512, 0. [Online]. Available:
  \url{https://doi.org/10.1177/0278364919846512}
\BIBentrySTDinterwordspacing

\bibitem{Deits2015EfficientMP}
R.~Deits and R.~Tedrake, ``Efficient mixed-integer planning for uavs in
  cluttered environments,'' \emph{2015 IEEE International Conference on
  Robotics and Automation (ICRA)}, pp. 42--49, 2015.

\bibitem{DBLP:journals/corr/abs-1811-10753}
\BIBentryALTinterwordspacing
W.~Sun, G.~Tang, and K.~Hauser, ``Fast {UAV} trajectory optimization using
  bilevel optimization with analytical gradients,'' \emph{CoRR}, vol.
  abs/1811.10753, 2018. [Online]. Available:
  \url{http://arxiv.org/abs/1811.10753}
\BIBentrySTDinterwordspacing

\end{thebibliography}

\end{document}